\documentclass[10pt,twocolumn,letterpaper]{article}

\usepackage{wacv}
\makeatletter
\@namedef{ver@everyshi.sty}{}
\makeatother
\usepackage{tikz}
\usepackage{times}
\usepackage{epsfig}
\usepackage{multicol}
\usepackage{multirow}
\usepackage{graphicx}
\usepackage{booktabs}
\usepackage{url}
\usepackage{amsmath}
\usepackage{latexsym}
\usepackage{amssymb}
\usepackage{float}
\usepackage{enumitem}
\usepackage{pgfplots}
\usepackage{cite}
\usepackage{subcaption}

\wacvfinalcopy 

\def\wacvPaperID{1074} 

\ifwacvfinal\fi
\begin{document}

\title{Learn like a Pathologist: Curriculum Learning by Annotator Agreement \\ for Histopathology Image Classification}

\author{Jerry Wei$^{1}$
, Arief Suriawinata$^{2}$
,  Bing Ren$^{2}$
,  Xiaoying Liu$^{2}$
,  Mikhail Lisovsky$^{2}$,\\  
Louis Vaickus$^{2}$,
Charles Brown$^{2}$
,  Michael Baker$^{2}$
,  Mustafa Nasir-Moin$^{1}$, \\
\vspace{1mm}
Naofumi Tomita$^{1}$, 
  Lorenzo Torresani$^{1}$
,  Jason Wei$^{1}$
,  Saeed Hassanpour$^{1\dagger}$ \\
$^{1}$Dartmouth College $^{2}$Dartmouth-Hitchcock Medical Center\\
$^\dagger$\texttt{saeed.hassanpour@dartmouth.edu}
}

\maketitle
\ifwacvfinal\thispagestyle{empty}\fi

\begin{abstract}
\vspace{-4.6mm}
Applying curriculum learning requires both a range of difficulty in data and a method for determining the difficulty of examples. 
In many tasks, however, satisfying these requirements can be a formidable challenge.

In this paper, we contend that histopathology image classification is a compelling use case for curriculum learning.
Based on the nature of histopathology images, a range of difficulty inherently exists among examples, and, since medical datasets are often labeled by multiple annotators, annotator agreement can be used as a natural proxy for the difficulty of a given example.
Hence, we propose a simple curriculum learning method that trains on progressively-harder images as determined by annotator agreement.

We evaluate our hypothesis on the challenging and clinically-important task of colorectal polyp classification.
Whereas vanilla training achieves an AUC of 83.7\% for this task, a model trained with our proposed curriculum learning approach achieves an AUC of 88.2\%, an improvement of 4.5\%. 
Our work aims to inspire researchers to think more creatively and rigorously when choosing contexts for applying curriculum learning.
\end{abstract}

\vspace{-6mm}
\section{Introduction}

\begin{figure}[ht]
    \centering
    \includegraphics[width=.9\linewidth]{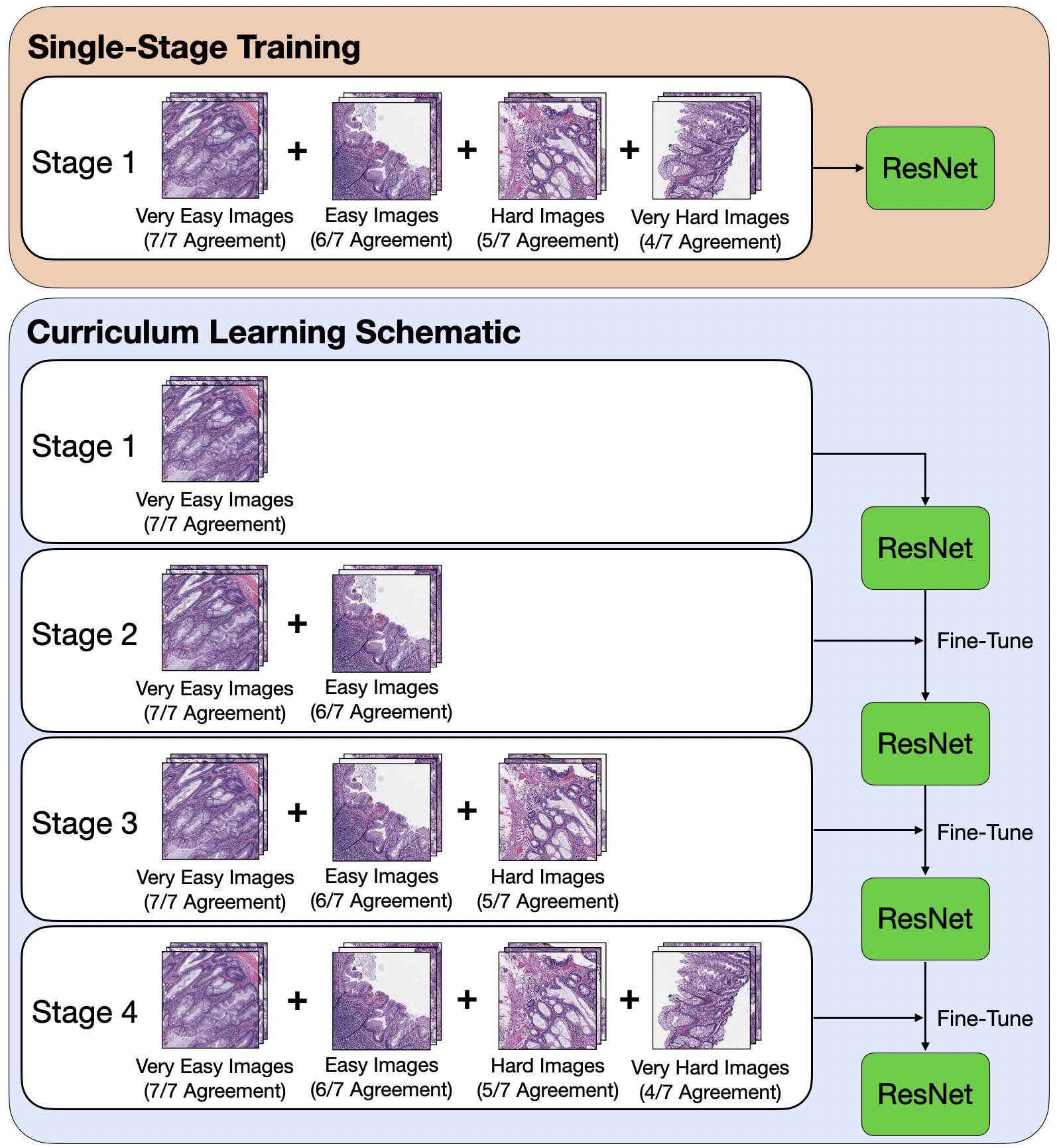}
    \vspace{-2.7mm}
    \caption{Our proposed curriculum learning by annotator agreement scheme for training a colorectal polyp classifier. The classifier first trains on easy images. Progressively harder images are gradually added in subsequent stages.}
    \vspace{-7mm}
    \label{fig:pull_figure}
\end{figure}

\vspace{-2mm}
Curriculum learning \cite{Bengio2009} is an elegant idea inspired by human learning that proposes that neural networks should be trained on examples in a specified order based on difficulty (typically easy to hard), as opposed to the random ordering that is currently common in practice. 
As such, curriculum learning requires both that there exists some range of difficulty among training examples and that we define a method for ranking examples.
In most cases, however, it is unclear whether a range of difficulty exists among the examples, and even when a range of difficulty exists, an ideal ranking function is rarely available. 
In this paper, we try to answer the question: are there tasks with domain-specific properties that are naturally appropriate for addressing these challenges of curriculum learning?

Interest in using deep learning to analyze histopathology images (stained tissues and cells that are typically manually examined under a microscope) has increased in recent years, with neural networks achieving pathologist-level performance on a variety of tasks \cite{Arvaniti2018,Bulten2020,Hekler2019,Shah2017,Strom2019,Wei2019,Zhang2019}. 
In this paper, we propose that histopathology image analysis is a suitable scenario for curriculum learning for two reasons.
First, due to the nature of histopathology images, we know that a range of difficulty in examples exists for many tasks. 
Second, medical image datasets typically have annotations from multiple clinicians---these annotations can be leveraged as a natural proxy for ranking example difficulty. 
Specifically, our paper makes the following contributions:
\begin{enumerate}[noitemsep,leftmargin=*]
    \item We contend that histopathology image classification is a natural scenario for applying curriculum learning, and we propose a curriculum learning approach that leverages annotator agreement as a proxy for difficulty.
    \item We evaluate our proposed approach on a colorectal polyp classification dataset, for which a baseline model achieved an AUC of 83.7\% and our best single-stage baseline achieved an AUC of 84.6\%. 
    When trained with curriculum learning, our model's AUC improves to 88.2\%, outperforming the average pathologist annotator on our test set in terms of Cohen's $\kappa$ \cite{Cohen1960}.  
\end{enumerate}
\vspace{-1.2mm}
The rest of our paper is outlined as follows. 
$\S$\ref{sec:curriculum-learning-intuitions} analyzes the challenges of curriculum learning and presents our intuitions on why histopathology image classification is a suitable context for curriculum learning. 
$\S$\ref{sec:histopathology-dataset} describes our task and dataset.
$\S$\ref{sec:formal-method} presents the main results of our proposed curriculum learning approach.
$\S$\ref{sec:proxies} compares the value of annotator agreement with two previously-proposed methods of scoring difficulty.
$\S$\ref{sec:path-performance} shows how the increase in AUC from curriculum learning translates to improvements in performance relative to pathologist performance.
$\S$\ref{sec:discussion} discusses the implications of our work. $\S$\ref{sec:further-related-work} puts our study in the context of prior work on curriculum learning for medical imaging and concludes our paper. 

\section{Curriculum Learning Intuitions}
\label{sec:curriculum-learning-intuitions}
\noindent \textbf{Curriculum learning.}
One of the earliest works demonstrating the benefit of curriculum learning \cite{Bengio2009} posits that learning occurs better when examples are not randomly presented but instead organized in a meaningful order that gradually shows more concepts and complexity.
Although the intuition behind this approach seems obvious in the context of human and animal learning, it is often unclear how to best apply this strategy for training neural networks.

As such, a diverse set of approaches has been explored in this area of research.
These approaches generally first score examples by difficulty and then train models using a schedule based on example difficulty, where easier examples are typically seen first and harder examples are seen later.
For instance, Bengio et al.'s original work \cite{Bengio2009} explored a noising-based curriculum for shape detection and a vocabulary-size based task for language modeling.
As popular recent examples, Weinshall et al.\ \cite{Weinshall2018} use the confidence of a pre-trained classifier as an estimator for difficulty; Korbar et al.\ \cite{Korbar2018} use a schedule with self-defined easy and hard examples for learning of audio-visual temporal synchronization; Ganesh and Corso\ \cite{Ganesh2020} propose to incrementally learn labels instead of learning difficult examples; and various teacher-student frameworks have been proposed in the context of curriculum learning \cite{Racaniere2020,Singh2018}.

\vspace{2.2mm} \noindent \textbf{Challenges of curriculum learning.} 
Despite the appeal of teaching machines to learn like humans, curriculum learning has been seen by some \cite{Weinshall2018} as mostly remaining in the fringes of machine learning research.
Based on the strategies of prior work, we broadly see two central challenges that arise when applying curriculum learning.

First, curriculum learning assumes that a range of easy and hard examples exists.
Although it could be argued that this is a true statement for any given dataset for at least some definition of easy and hard, the distribution of example difficulties likely varies based on the nature of the task and the dataset. 
Since the added value of curriculum learning comes from utilizing the varying degrees of difficulty in a task, tasks with a smaller range of example difficulty are less conducive to effective curriculum learning. 
Empirically, Weinshall et al., present some evidence related to this claim, showing that curriculum learning had a larger improvement compared with regular training when applied to tasks that were more difficult and likely included challenging examples (e.g., distinguishing small mammals in CIFAR-100), than when applied to tasks that were easier and did not have examples that were difficult to classify (e.g., discriminating between 5 well-separated classes in CIFAR-100) \cite{Weinshall2018}. 

Second, a curriculum learning approach must somehow categorize examples as easy or hard.
Prior work has tried to address this challenge in many ways, including trying to discover inherent patterns in the data \cite{Weinshall2018}, using hand-picked heuristics \cite{Tsvetkov2016}, and creating custom training progressions \cite{Korbar2018}.
Though sometimes effective, these methods can be difficult to implement, and it is often unclear whether an approach that works on one dataset will also work on another.
Indeed, scoring images by difficulty is often the core problem addressed in many curriculum learning papers.

\vspace{2.2mm} \noindent \textbf{Our intuitions.} 
We contend that histopathology image classification is an important task that naturally addresses the challenges above---and could benefit from curriculum learning---based on the following two observations:
\begin{itemize}[leftmargin=*]
    \item \textbf{A range of example difficulty exists in many histopathology image classification tasks.} 
    We believe this to be true for several domain-specific reasons.
    (1) Because pathological disease develops over time, there is a progression from normal tissue to diseased tissue. Since many diseases are classified into discrete classes, there must be some points in this progression that lie on the margins of two classes and are therefore hard to diagnose.
    (2) Pathology residents learn to read images by first studying classic examples of diseases and then learning to diagnose more-challenging cases over time, implying that human instructors acknowledge some notion of easy and hard examples \cite{Mais2013}.
    (3) Inter-annotator agreement is moderate or low on many disease classification tasks \cite{Wei2019,Wei2020}, suggesting that some images are hard to classify.
    \textit{Knowing that a curriculum exists is the first step to applying curriculum learning.}
    \item \textbf{Image-level annotator agreement can be leveraged as a proxy for example difficulty.} 
    As medical image datasets are commonly annotated by several trained clinicians, we can leverage the extent of agreement for each image as a proxy for the difficulty of that image. 
    By definition, images with high agreement are easy to classify, as everyone agrees on them, and images with lower agreement are harder to classify.
    If these human notions of difficulty translate to a helpful curriculum for training neural networks, then many tasks with annotator agreement data already contain a curriculum that can be used to improve model performance. \cite{Chilamkurthy2018,Coudray2017,Bejnordi2017,Esteva2017,Ghorbani2019,Gulshan2016,Irvin2019,Kanavati2020,Korbar2017,Sertel2008,Wang2019,Wei2020,Zhou2019}.
\end{itemize}

\section{Histopathology Dataset \protect \footnote{We plan to make our dataset and annotations publicly available to facilitate further research.}}
\label{sec:histopathology-dataset}

In this paper, we focus on the task of colorectal polyp classification, a challenging and clinically-important task in pathology. 
As shown in Table \ref{tab:train_test_split}, our dataset contains 3,152 images in total, each annotated with a binary label of either hyperplastic polyp (HP) or sessile serrated adenoma (SSA). 

\vspace{2.2mm} \noindent \textbf{Colorectal polyp classification task.} 
Colonoscopy is a common screening program in the United States \cite{Rex2017}, and so classification of colorectal polyps (growths inside the colon lining that can lead to colonic cancer if left untreated) is one of the highest-volume tasks in pathology.
Our task focuses on the clinically-important binary distinction between hyperplastic polyps (HPs) and sessile serrated adenomas (SSAs), a challenging problem \cite{Abdeljawad2015,Farris2008,Glatz2007,Khalid2009,Wong2009}.
Pathologically, SSAs are characterized by broad-based crypts, often with complex structure and heavy serration \cite{Gurudu2010}.

\vspace{2.2mm} \noindent \textbf{Data collection and annotation.}
For our data collection, we scanned 328 Formalin Fixed Paraffin-Embedded (FFPE) whole-slide images of colorectal polyps, which were originally diagnosed as either hyperplastic polyps (HPs) or sessile serrated adenomas (SSAs), from patients at our tertiary medical institution.
From these 328 whole-slide images, we then extracted 3,152 patches (image portions of size 224 $\times$ 224 pixels) representing diagnostically-relevant regions of interest for HPs or SSAs.
The seven practicing board-certified gastrointestinal pathologists at our tertiary institution then independently labeled each of the 3,152 images in our dataset as either HP or SSA.

\vspace{2.2mm} \noindent \textbf{Train-test split and gold-standard labels.}
Images were split randomly by whole-slide such that images from the same whole-slide either all went into the training set or all went into the testing set.
As shown in Table \ref{tab:train_test_split}, we used a training set of 2,175 images ($\sim$70\% of images) and a testing set of 977 images ($\sim$30\% of images).
In our testing set, we use the majority vote of labels as the gold-standard label, a common choice in the literature \cite{Chilamkurthy2018,Gulshan2016,Irvin2019,Kanavati2020,Korbar2017,Sertel2008,Wang2019,Wei2019Celiac,Wei2020Difficulty,Zhou2019}. 

\vspace{4mm}
\begin{table}[t!]
    \small
    \centering
    \begin{tabular}{l c c c}
        \toprule
        & Train & Test & Total\\
        \midrule
         HP & 1,545 & 617 & 2,162\\
         SSA & 630 & 360 & 990\\
         \midrule
         Total & 2,175 & 977 & 3,152\\
         \bottomrule
    \end{tabular}
    \vspace{-2mm}
    \caption{Number of images in our dataset.}
    \label{tab:train_test_split}
    \vspace{3mm}
\end{table}
\vspace{10mm}
\begin{table}[t!]
\setlength{\tabcolsep}{3pt}
    \centering
    \begin{tabular}{l |  c c c c c c c}
        \toprule
         & \hspace{1.5mm}A1\hspace{1mm} & A2 & A3 & A4 & A5 & A6 & A7 \\
        \midrule
        \hspace{1mm}A1\hspace{1mm} & - & 65.7 & 90.1 & 82.0 & 71.5 & 90.7 & 63.6\\
        \hspace{1mm}A2\hspace{1mm} & - & - & 64.2 & 76.0 & 76.1 & 65.8 & 60.8\\
        \hspace{1mm}A3\hspace{1mm} & - & - & - & 80.1 & 69.3 & 90.8 & 62.3\\
        \hspace{1mm}A4\hspace{1mm} & - & - & - & - & 79.9 & 81.9 & 64.1\\
        \hspace{1mm}A5\hspace{1mm} & - & - & - & - & - & 70.7 & 61.5\\
        \hspace{1mm}A6\hspace{1mm} & - & - & - & - & - & - & 62.9\\
        \hspace{1mm}A7\hspace{1mm} & - & - & - & - & - & - & -\\
        \bottomrule
    \end{tabular}
    \vspace{-2mm}
    \caption{Pairwise annotator agreement (\%) for our seven annotators (indexed as A1, A2, A3, A4, A5, A6, and A7).}
    \label{tab:annotator_pair_agreement}
    \vspace{3mm}
\end{table}
\vspace{5mm}
\begin{figure}[t!]
    \centering
    \includegraphics[width=\linewidth]{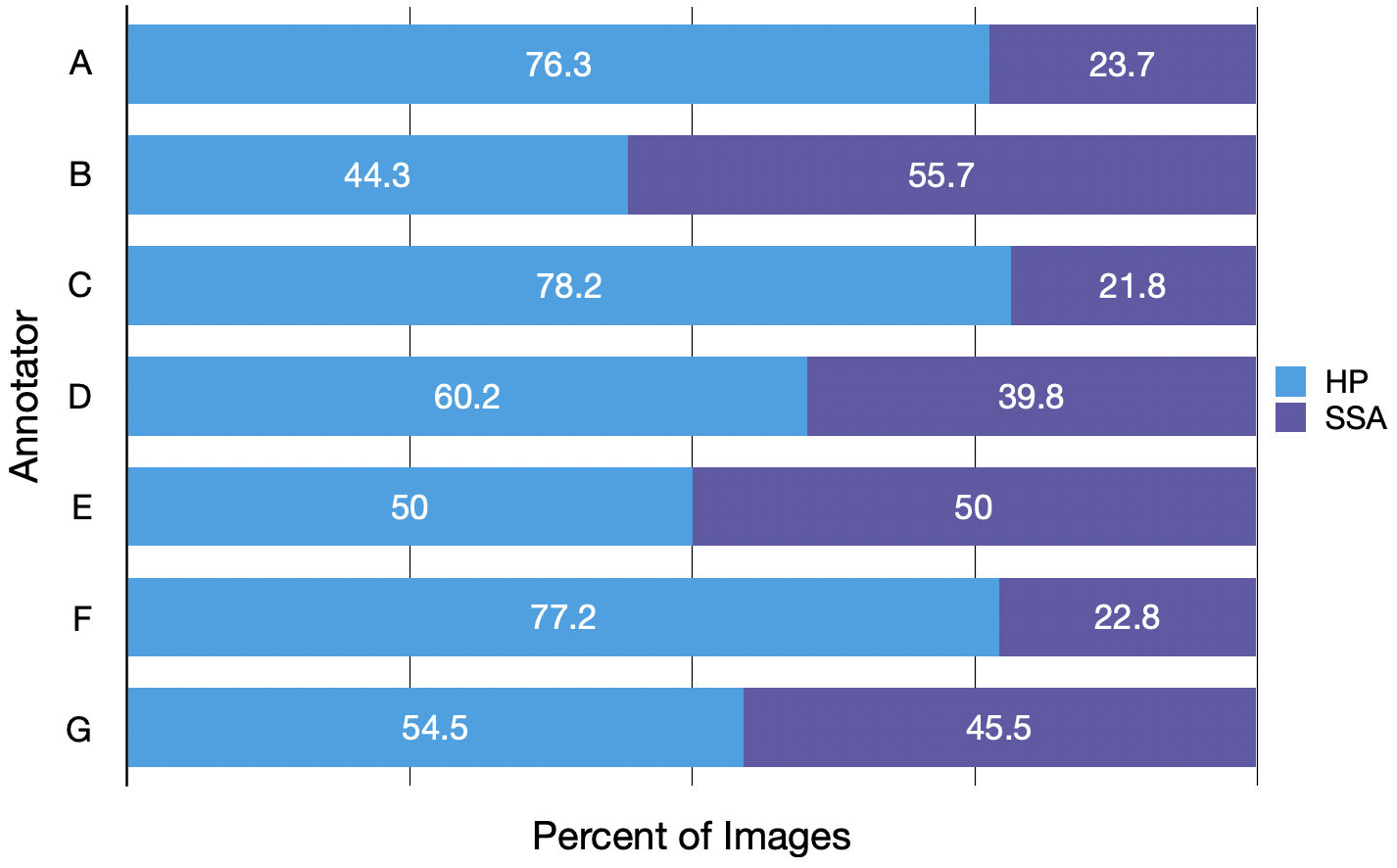}
    \vspace{-7mm}
    \caption{Distribution of class labels for each annotator.}
    \label{fig:annotator_class_distribution}
    \vspace{3mm}
\end{figure}
\vspace{5mm}
\begin{figure}[h!]
    \centering
    \includegraphics[width=0.95\linewidth]{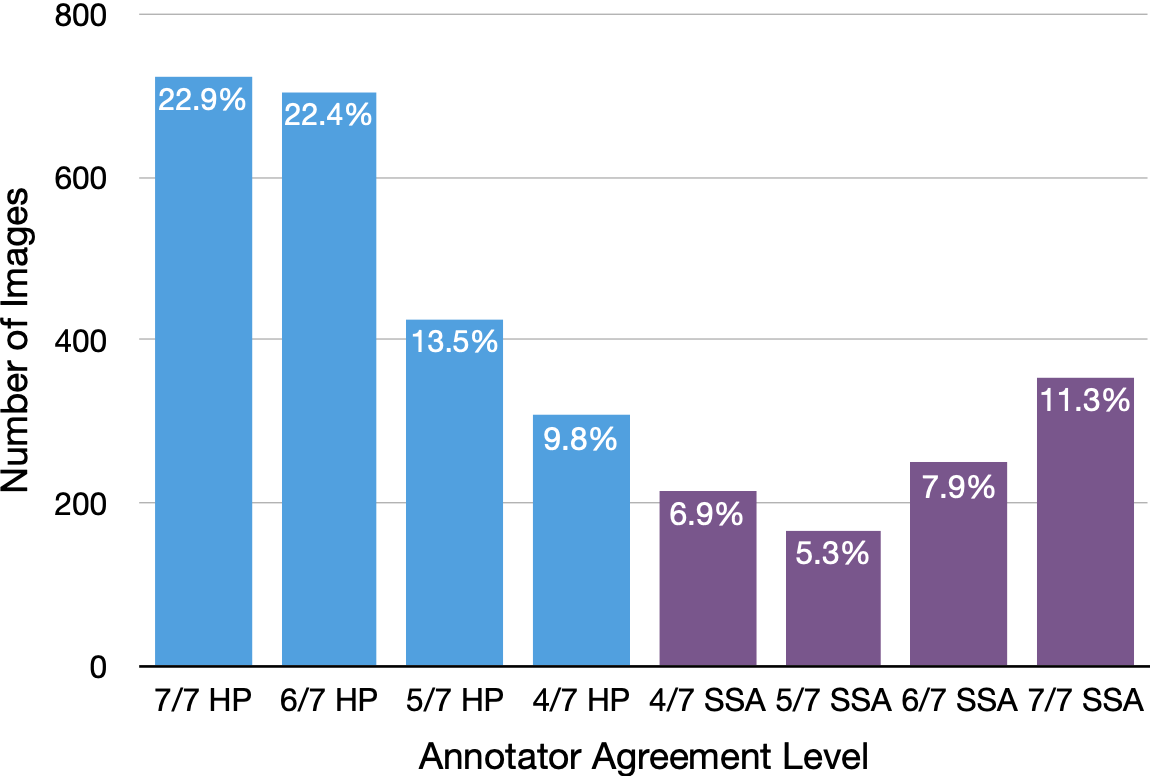}
    \vspace{-2mm}
    \caption{Distribution of annotator agreement levels for images in our dataset.}
    \label{fig:annotator_agreement_distribution}
\end{figure}
\vspace{3mm}
\begin{figure*}[t!]
    \centering
    \includegraphics[width=0.7\linewidth]{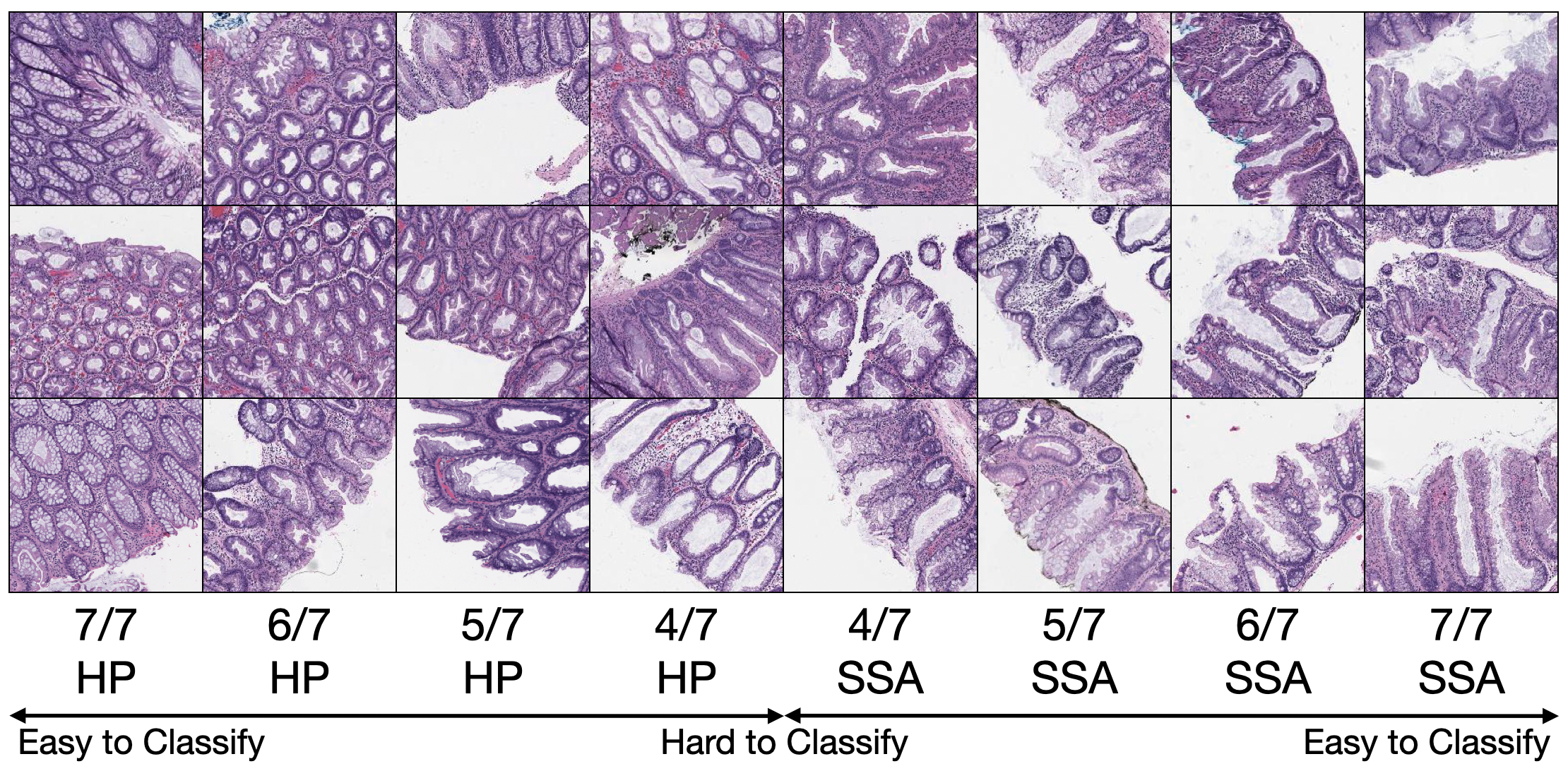}
    \vspace{-1.7mm}
    \caption{Example images for each level of annotator agreement.}
    \vspace{-2mm}
    \label{fig:agreement_examples}
    
\end{figure*}

\vspace{2.2mm} \noindent \textbf{Dataset Statistics.}
To help readers get a better sense of our dataset, in Tables \ref{tab:train_test_split} and \ref{tab:annotator_pair_agreement} and in Figures \ref{fig:annotator_class_distribution} and \ref{fig:annotator_agreement_distribution}, we show several analyses of our dataset.
To summarize, our dataset contains 2,162 images with a gold-standard label of HP and 990 images with a gold-standard label of SSA, with 64.5\% of images having at 6/7 or 7/7 annotator agreement and 35.5\% of images with annotator agreement of 4/7 or 5/7.
Figure \ref{fig:agreement_examples} shows examples for each level of agreement.
On average, each pair of pathologists had an average agreement of 72.9\%, and each pathologist agreed with the majority vote 83.2\% of the time, indicating that our task has non-neglible disagreement, even among pathologist annotators who all specialize in gastroenterology.

\begin{table*}[t!]
    \centering
    \small
    \begin{tabular}{l | c | c c c c | c | c }
    \toprule
    & & \multicolumn{5}{c|}{AUC (\%) on test set stratified by annotator agreement} \\
    & Stage & Very Easy & Easy & Hard & Very Hard & \underline{Overall} & $\Delta^{+}$ \\
    \midrule
    \textbf{Single-Stage Training} & & & & & & &\\
    Vanilla Baseline: All Images & - & 93.8 $\pm$ 0.5 & 88.7 $\pm$ 1.3 & 76.2 $\pm$ 0.9 & 60.7 $\pm$ 2.1 & 83.7 $\pm$ 1.0 & - \\
    Very Easy Images Only & - & 94.8 $\pm$ 0.8 & 87.7 $\pm$ 1.1 & 61.7 $\pm$ 1.9 & 56.2 $\pm$ 1.6 & 80.2 $\pm$ 1.1 &  \\
    Easy Images Only & - & 93.7 $\pm$ 0.9 & 88.6 $\pm$ 0.7 & 73.8 $\pm$ 1.8 & 56.4 $\pm$ 1.7 & 82.7 $\pm$ 0.8 &  \\
    Very Easy + Easy Images & - & 96.1 $\pm$ 0.6 & 90.2 $\pm$ 1.2 & 72.1 $\pm$ 1.7 & 58.0 $\pm$ 1.9 & 84.6 $\pm$ 0.8 & * \\
    Very Easy + Easy + Hard Images & - & 94.7 $\pm$ 0.8 & 88.8 $\pm$ 1.2 & 76.0 $\pm$ 1.3 & 60.2 $\pm$ 1.9 & 84.2 $\pm$ 0.8 &   \\
    \midrule
    \multicolumn{8}{l}{\textbf{Curriculum Learning - Annotator Agreement (Ours)}}\\
    Very easy images & 1 & 94.8 $\pm$ 0.8 & 87.7 $\pm$ 1.1 & 61.7 $\pm$ 1.9 & 56.2 $\pm$ 1.6 & 80.2 $\pm$ 1.1 &  \\
    \hspace{2mm} then (very easy + easy) & 2 & 96.2 $\pm$ 0.6 & 91.3 $\pm$ 1.1 & 74.1 $\pm$ 1.4 & 58.8 $\pm$ 2.2 & 85.5 $\pm$ 0.9 & ***\\
    \hspace{4mm} then (very easy + easy + hard) & 3 & \textbf{96.7} $\pm$ \textbf{0.5} & \textbf{94.3} $\pm$ \textbf{0.5} & \textbf{78.9} $\pm$ \textbf{1.2} & 64.2 $\pm$ 2.0 & \textbf{88.2} $\pm$ \textbf{0.6} & ***\\
    \hspace{6mm} then (very easy + easy + hard + very hard) & 4 & 96.1 $\pm$ 0.6 & 93.2 $\pm$ 1.2 & 78.3 $\pm$ 1.6 & \textbf{64.5} $\pm$ \textbf{1.4} & 87.1 $\pm$ 0.9 & ***\\
    \midrule
    \multicolumn{8}{l}{\textbf{ANTI-Curriculum Learning - Annotator Agreement}}\\
    Very hard images & 1 & 66.2 $\pm$ 4.5 & 60.9 $\pm$ 4.3 & 60.5 $\pm$ 4.2 & 55.8 $\pm$ 2.4 & 59.6 $\pm$ 3.2 &  \\
    \hspace{2mm} then (very hard + hard) & 2 & 71.5 $\pm$ 4.5 & 67.6 $\pm$ 5.9 & 65.3 $\pm$ 2.6 & 56.3 $\pm$ 3.4 & 65.7 $\pm$ 4.5 &  \\
    \hspace{4mm} then (very hard + hard + easy) & 3 & 89.6 $\pm$ 1.5 & 86.3 $\pm$ 1.1 & 73.2 $\pm$ 3.8 & 60.0 $\pm$ 2.7 & 80.0 $\pm$ 1.1 &  \\
    \hspace{6mm} then (very hard + hard + easy + very easy ) & 4 & 93.7 $\pm$ 0.8 & 88.3 $\pm$ 1.1 & 76.8 $\pm$ 1.4 & 61.3 $\pm$ 1.5 & 83.6 $\pm$ 0.7 &  \\
    \midrule
    \multicolumn{8}{l}{\textbf{Curriculum Learning - Direct Annotation}}\\
    Very easy images & 1 & 90.8 $\pm$ 1.6 & 88.6 $\pm$ 1.3 & 74.7 $\pm$ 3.2 & 60.1 $\pm$ 1.8 & 82.5 $\pm$ 1.3 &  \\
    \hspace{2mm} then (very easy + easy) & 2 & 93.0 $\pm$ 0.8 & 88.3 $\pm$ 0.6 & 77.1 $\pm$ 1.5 & 60.2 $\pm$ 1.5 & 83.3 $\pm$ 0.5 &  \\
    \hspace{4mm} then (very easy + easy + hard) & 3 & 93.4 $\pm$ 0.7 & 88.4 $\pm$ 0.8 & 77.1 $\pm$ 1.0 & 59.9 $\pm$ 2.0 & 83.5 $\pm$ 0.7 &  \\
    \hspace{6mm} then (very easy + easy + hard + very hard ) & 4 & 93.2 $\pm$ 0.8 & 88.1 $\pm$ 1.0 & 77.3 $\pm$ 1.4 & 60.6 $\pm$ 2.2 & 83.3 $\pm$ 0.6 &  \\
    \midrule
    \multicolumn{8}{l}{\textbf{Curriculum Learning - Control (Random)}}\\
    Very easy images & 1 & 89.8 $\pm$ 1.2 & 88.7 $\pm$ 0.9 & 68.9 $\pm$ 2.4 & 57.8 $\pm$ 1.8 & 80.3 $\pm$ 1.2 &  \\
    \hspace{2mm} then (very easy + easy) & 2 & 92.1 $\pm$ 0.8 & 88.2 $\pm$ 0.9 & 76.2 $\pm$ 1.4 & 59.4 $\pm$ 1.5 & 82.6 $\pm$ 0.5 &  \\
    \hspace{4mm} then (very easy + easy + hard) & 3 & 93.2 $\pm$ 0.6 & 89.2 $\pm$ 0.8 & 76.3 $\pm$ 1.3 & 58.7 $\pm$ 1.6 & 83.4 $\pm$ 0.6 &  \\
    \hspace{6mm} then (very easy + easy + hard + very hard ) & 4 & 93.6 $\pm$ 0.7 & 89.2 $\pm$ 1.4 & 76.8 $\pm$ 1.6 & 59.8 $\pm$ 2.4 & 83.7 $\pm$ 0.9 &  \\
    \bottomrule
    \end{tabular}
    \vspace{-1.0mm}
    \caption{
    Histopathology image classification model trained using a curriculum learning framework outperforms single-stage training baselines by an AUC of 3.6--4.5\%. 
    Image difficulty is determined by annotator agreement in four discrete categories: \textit{very easy} (7/7 annotator agreement), \textit{easy} (6/7 agreement), \textit{hard} (5/7 agreement), and \textit{very hard} (4/7 agreement).
    $\Delta^{+}$ indicates the level of statistical significance in improvement over the vanilla baseline of training with all images: * indicates $p \leq 0.05$; *** indicates $p \leq 0.001$.
    Means and standard deviations shown are for 20 random seeds.
    }
    \label{tab:main_table}
    \vspace{-2.3mm}
\end{table*}

\section{Curriculum learning: annotator agreement}
\label{sec:formal-method}

We propose a curriculum learning framework that leverages annotator agreement to rank images by difficulty. 
Specifically, we define images with high annotator agreement to be easy and images with low annotator agreement to be hard. 
For our dataset, which was labeled by seven annotators, we partition our images into four discrete levels of difficulty: \textit{very easy} (7/7 agreement among annotators), \textit{easy} (6/7 agreement among annotators), \textit{hard} (5/7 agreement among annotators), and \textit{very hard} (4/7 agreement among annotators).
An overview schematic of this setup can be seen in Figure \ref{fig:pull_figure}.
For our training schedule, we train our network on progressively harder images in four stages:
\begin{itemize}[noitemsep,leftmargin=*]
    \item Stage 1: \textit{Very easy} images only
    \item Stage 2: \textit{Very easy} + \textit{easy} images
    \item Stage 3: \textit{Very easy} + \textit{easy} + \textit{hard} images
    \item Stage 4: \textit{Very easy} + \textit{easy} + \textit{hard} + \textit{very hard} images
\end{itemize}
At each stage, we make sure to include images from the previous stages to prevent catastrophic forgetting \cite{French1999}.
Though our current model uses four levels of difficulty and four stages of training, our general framework could be used in any scenario where annotator agreement data is available.

\vspace{2.2mm} \noindent \textbf{Experimental Setup.} 
For our model, we use ResNet-18, a common choice for classifying histopathology images. 
Specifically, we follow the DeepSlide repository \cite{Wei2019} for histopathology image classification, training our model for 50 epochs (well past convergence) using stochastic data augmentation with the Adam optimizer \cite{Kingma2014}, initial learning rate of $1\times10^{-3}$, and learning rate decay factor of 0.91. 

For more-robust evaluation, for each model we consider the five highest AUCs on the test set, which are evaluated at every epoch. 
We report the mean and standard deviation of these values calculated over 20 different random seeds.

\vspace{2.2mm} \noindent \textbf{Baselines.}
Our primary baseline is the vanilla-training model where all training images are given a label determined by the majority vote of annotator labels, a common gold-standard in the literature \cite{Chilamkurthy2018,Gulshan2016,Irvin2019,Kanavati2020,Korbar2017,Sertel2008,Wang2019,Wei2019Celiac,Wei2020Difficulty,Zhou2019}. 
We also explore variations of single-stage training in which only certain images, selected based on annotator agreement, are used for training.
For all experiments, our test set is fixed and contains images from all levels of difficulty, although stratified analyses are also presented.

As shown in the first block of Table \ref{tab:main_table}, our vanilla baseline that uses all images achieves an AUC of 83.7\%. 
Interestingly, the network trained on only \textit{very easy} and \textit{easy} images achieved an overall AUC of 84.6\%, an almost 1\% improvement over the vanilla baseline.
As shown by the performances on the test set when stratified by difficulty, this network  trained on \textit{very easy} and \textit{easy} images does better on \textit{very easy} and \textit{easy} images in the test set while doing worse on \textit{hard} and \textit{very hard} images, leading to a higher overall AUC because there are more \textit{very easy} and \textit{easy} images than \textit{hard} and \textit{very hard} image in the testing set.

\vspace{2.2mm} \noindent \textbf{Annotator agreement-based curriculum learning.}
The second block of Table \ref{tab:main_table} shows the results of our proposed curriculum learning scheme at each stage of training.
In this curriculum learning scheme, the model already outperforms all single-stage models at the second stage, achieving an AUC of 85.5\%, and at the third stage, achieves an AUC of 88.2\%, the highest of any model we train.
Moreover, this model also achieved the highest performance when stratified by \textit{very easy}, \textit{easy}, and \textit{hard} images in the test set, outperforming earlier stages that trained only on \textit{very easy} and \textit{easy} examples.
These results suggest that training on harder images in a curriculum framework not only improves performance on hard images, but also improves performance on easy images, a finding consistent with Korbar et al.\ \cite{Korbar2018}.

Perhaps strikingly, model performance actually decreases in the fourth stage of training that includes \textit{very hard} examples, as performance on \textit{very hard} images in the test set increases but performance on other images in the test set decreases.
One explanation for this slight dip in performance is that \textit{very hard} images, which have only 4/7 pathologist agreement, could be too challenging to analyze accurately (even for expert humans), so their features might not be beneficial for training machine learning models either.

In terms of statistical significance, our curriculum learning model at the second, third, and fourth stages outperforms the vanilla-training model with $p\leq0.001$, based on a two-sample \textit{t}-test for means. 

\vspace{1.3mm} \noindent \textbf{Anti-curriculum learning.} 
To further validate that the improvement in model performance is indeed a result of the intentionally selected images at each stage, we train a model using an \textit{anti-curriculum} scheme, which reverses the learning schedule (i.e., the model first trains on hard images and then trains on progressively easier images). 
As shown in the third block in Table \ref{tab:main_table}, no models trained using the anti-curriculum framework outperform the vanilla baseline.

\begin{figure}[t]
    \centering
    \includegraphics[width=\linewidth]{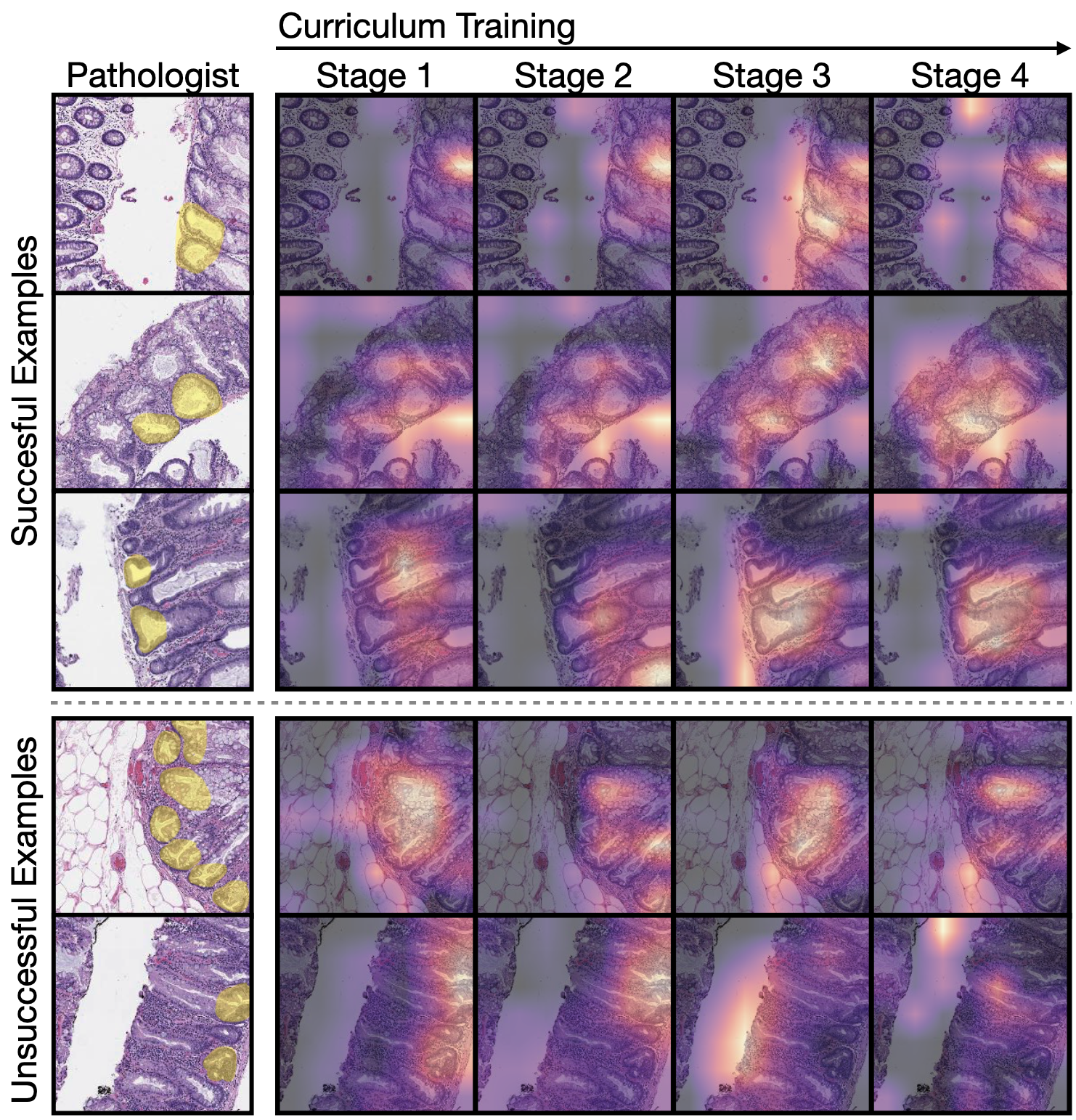}
    \vspace{-6mm}
    \caption{GradCAM visualization of images where curriculum learning was successful (top) and unsuccessful (bottom). Regions of interest are highlighted in yellow. For successful images, pathologists marked Stage 3 (our best-performing model) as the best representation of the area that they would look at to make a diagnosis. For unsuccessful images, pathologists marked Stage 1 as the best.}
    \label{fig:heatmap_vis}
    \vspace{-5.2mm}
\end{figure}
\vspace{1.3mm} \noindent \textbf{Visualization.} For a qualitative examination of how the model changes throughout training, we compute GradCAM heatmaps \cite{Selvaraju2016} to visualize the model's predictions at each of the four stages of curriculum learning. 
In Figure \ref{fig:heatmap_vis}, we show examples of SSA images where curriculum learning was both successful and unsuccessful, as subjectively examined by our pathologists.
In the successful examples, the model seemed to focus on broad-based crypts (a defining characteristic of SSAs) more heavily in Stage 3 of curriculum training, our best-performing model.
In the unsuccessful examples, on the other hand, the model seemed to focus on  broad-based crypts more heavily in earlier stages. 

\vspace{1.3mm} \noindent \textbf{Other curriculum baselines.} 
As a further baseline, we also asked a pathologist to directly score the difficulty of each image on a scale from 1-4, with 1 as very easy to classify and 4 as very hard to classify. 
We also ran a four-stage experiment using this difficulty measure, as shown in the fourth block of Table \ref{tab:main_table}, but find that curriculum learning here does not improve performance, possibly because the manual difficulty scores from a single pathologist are too subjective. 
Moreover, we tested a control curriculum with the same training scheme as the annotator agreement experiment, except images in each stage were selected randomly (fifth block of Table \ref{tab:main_table}).
As expected, this control curriculum performed about the same as our vanilla baseline.

\section{Annotator agreement vs model confidence}
\label{sec:proxies}
This section compares various proxies for difficulty in terms of their usefulness for curriculum learning.
Whereas our model so far has used annotator agreement as a proxy for example difficulty, prior work has proposed that the output confidence of a machine learning classifier can also be a proxy for example difficulty \cite{Hacohen19,Weinshall2018}. 
First, we perform a sanity check to see whether classifier output confidence correlates with annotator agreement.
As we might expect, they do---predicted confidence distributions of a model pre-trained on ImageNet and fine-tuned on our dataset appeared substantially different for different levels of annotator agreement (Figure \ref{fig:agreement_distributions} in the Supplementary Materials).

We conduct a simple ablation study to compare annotator agreement and model output confidence as proxies for image difficulty in curriculum learning.
For simplicity, we use a two-step curriculum learning scheme---where training is done in one stage containing only easy images and a following stage containing a mixture of easy and hard images---and use a single-step pacing schedule \cite{Hacohen19}.

We evaluate the following three proxies for difficulty:
\begin{enumerate}[noitemsep,leftmargin=*]
    \item \textbf{Self-taught scoring}, where the classifier with randomly initialized weights is pre-trained on our dataset, and output confidences are used to sort examples by difficulty.
    \item \textbf{Transfer learning}, where a classifier pre-trained on ImageNet is fine-tuned on our dataset, and output confidences are used to sort examples by difficulty.
    \item \textbf{Coarse annotator agreement}, a simplified version of our curriculum learning scheme above, where images are divided into two categories of either easy (6/7 or 7/7 agreement) or hard (4/7 or 5/7 agreement), instead of the four categories used in $\S$\ref{sec:formal-method}.
\end{enumerate}
The self-taught scoring and transfer learning proxies assign each image with a confidence score: images with confidence score greater than a threshold $\tau$ are classified as easy and images with confidence score less than $\tau$ are classified as hard.
We choose $\tau$ such that the proportion of easy and hard images was the same as the natural distribution of easy and hard images from coarse annotator agreement.
Then, a new classifier is trained in two stages: (1) easy images only, and (2) easy images and hard images combined at various ratios.
Our coarse annotator agreement method in this section follows this same training scheme of two stages. 

Figure \ref{fig:transfer-learning} shows the results for the various ratios of hard images we used in the second stage of training.
Without making any general claims, we see that on our dataset, annotator agreement appears to be a more useful proxy for difficulty than model output confidence.
One possible explanation for this result is that although the transfer learning and self-taught scoring approaches work (marginally, in our case), much of the information provided by the pre-trained classifier is shared with the resulting classifier (i.e., much of the added value of the pre-trained classifier can already be discovered by the resulting classifier), whereas using annotator agreement as a proxy provides new information about difficulty that the model would not have had access to otherwise.
Moreover, we find that including a greater proportion of easy images than hard images in the second stage of training is important for preventing catastrophic forgetting, a finding consistent with prior work \cite{Korbar2018,Weinshall2018,Hacohen19}.
Using only hard images in the second stage was especially problematic for the self-taught scoring and transfer learning models, possibly because the images most challenging for these proxy models to learn will also be challenging for the learner model to optimize.  

\pgfplotsset{width=8.5cm,height=6.4cm,compat=1.9}
    \vspace{-2mm}
\begin{figure}[t!]
\setlength{\abovecaptionskip}{0pt}
\begin{centering}
\begin{tikzpicture}
\begin{axis}[
    xlabel={Hard Image Ratio in Stage 2 (\%)},
    ylabel={AUC (\%)},
    xmin=23, xmax=103,
    ymin=51, ymax=87,
    xtick={25, 33, 50, 75, 100},
    ytick={55, 60, 65, 70, 75, 80, 85},
    legend pos=south west,
    ymajorgrids=true,
    xmajorgrids=true,
    grid style=dotted,
    ylabel style={yshift=-0.2cm},
]
\addplot[
    color=violet,
    mark=triangle,
    mark size=0pt,
    dashed,
    thick,
    ]
    coordinates {
    (23,    83.9)
    (103,	83.9)
    };
    \addlegendentry{Single-Stage Baseline}
\addplot[
    color=blue,
    mark=square,
    mark size=2pt,
    ]
    coordinates {
    (25,    84.9)
    (33,    85.2)
    (50,    82.9)
    (75,    81.0)
    (100,   78.7)
    };
    \addlegendentry{Annotator Agreement (Ours)}
\addplot[
    color=red,
    mark=o,
    mark size=2pt,
    ]
    coordinates {
    (25,    83.9)
    (33,    84.1)
    (50,    81.9)
    (75,    79.1)
    (100,   68.6)
    };
    \addlegendentry{Self-Taught Scoring}
\addplot[
    color=black,
    mark=triangle,
    mark size=3pt,
    ]
    coordinates {
    (25,    84.5)
    (33,    84.6)
    (50,    82.3)
    (75,    77.6)
    (100,   53.7)
    };
    \addlegendentry{Transfer Learning}
\end{axis}
\end{tikzpicture}
\caption{
Performance of curriculum learning models that use proxies of transfer learning \cite{Hacohen19}, self-taught scoring \cite{Hacohen19}, and annotator agreement (ours).
For all tested ratios of hard images in the second stage of training, curriculum learning by annotator agreement outperforms transfer learning and self-taught scoring.
}
\vspace{-4mm}
\label{fig:transfer-learning}
\end{centering}
\end{figure}
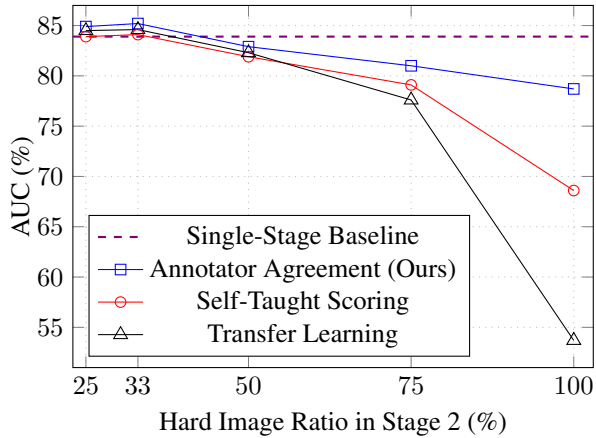

\section{Comparison with Pathologist Performance}
\label{sec:path-performance}

For more context on the significance of the improvement that curriculum learning brings to model performance, in this section, we compare the performance of our models with that of pathologists. 
Specifically, we frame the predictions of each model as the annotations of an additional pathologist, and we compare these predictions with the annotations of actual pathologists in terms of Cohen's $\kappa$ \cite{Cohen1960}, a common measure of inter-annotator agreement.

For our models, we select the best-performing curriculum learning model (Stage 3) and compare it with the vanilla-baseline model that was trained on all images in a single stage.
As our models output a continuous distribution of probabilities for HP and SSA, we evaluate each model at several different confidence thresholds (a lower threshold results in higher recall, whereas a higher threshold results in higher precision). 
For average pathologist performance, we compute the Cohen's $\kappa$ between all pairs of pathologists, and for each of the seven pathologists we show the mean Cohen's $\kappa$ of all six pairs involving that pathologist. 

Figure \ref{fig:cohens-kappa} shows these results comparing our models with both individual pathologists and the average of all pathologists.
We see that there is a wide range of Cohen's $\kappa$ for individual pathologists---the mean of our pathologists' Cohen's $\kappa$ scores was 0.450, which is in the \textit{moderate} range of 0.41-0.60 \cite{McHugh2012} (a similar study found a Cohen's $\kappa$ of 0.380 found among four pathologists \cite{Wong2009}).
Our curriculum learning model (AUC = 88.2\%) outperforms the pathologist average for multiple thresholds and the baseline model (AUC = 83.7\%) for all thresholds.
In particular, adding a curriculum schedule increases performance from the baseline's maximum $\kappa$\hspace{1mm}=\hspace{1mm}$0.384$ to the curriculum learning model's maximum $\kappa$\hspace{1mm}=\hspace{1mm}$0.473$, the difference needed to outperform the pathologists' average ($\kappa$\hspace{1mm}=\hspace{1mm}$0.450$). 

\pgfplotsset{width=8cm,height=6cm,compat=1.9}
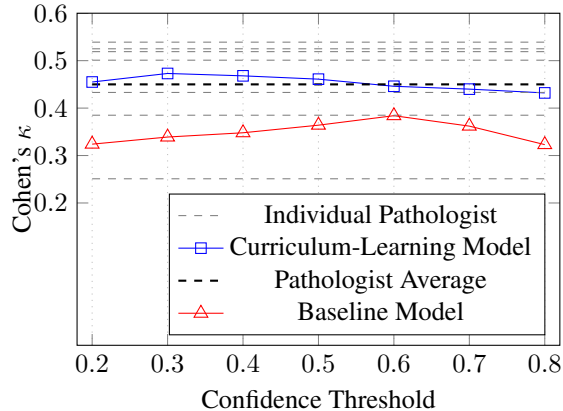
\begin{figure}[t]
\setlength{\abovecaptionskip}{0pt}
\begin{centering}
\begin{tikzpicture}
\begin{axis}[
    xlabel={Confidence Threshold},
    ylabel={Cohen's $\kappa$},
    xmin=0.18, xmax=0.82,
    ymin=-.10, ymax=.60,
    xtick={.2, .3, .4, .5, .6, .7, .8},
    ytick={.20, .30, .40, .50, .60},
    legend pos=south east,
    ymajorgrids=false,
    xmajorgrids=true,
    grid style=dotted,
    ylabel style={yshift=-0.2cm},
]
\addplot[
    color=gray,
    mark=triangle,
    mark size=0pt,
    dashed,
    ]
    coordinates {
    (.2,    .525)
    (.8,	.525)
    };
    \addlegendentry{Individual Pathologist}
\addplot[
    color=blue,
    mark=square,
    mark size=2pt,
    ]
    coordinates {
    (.2,    .455)
    (.3,    .473)
    (.4,    .468)
    (.5,    .461)
    (.6,    .446)
    (.7,    .440)
    (.8,    .432)
    };
    \addlegendentry{Curriculum-Learning Model}
\addplot[
    color=black,
    mark=triangle,
    mark size=0pt,
    dashed,
    thick,
    ]
    coordinates {
    (.2,    .45)
    (.8,	.45)
    };
    \addlegendentry{Pathologist Average}
\addplot[
    color=red,
    mark=triangle,
    mark size=3pt,
    ]
    coordinates {
    (.2,    .324)
    (.3,    .339)
    (.4,    .348)
    (.5,    .364)
    (.6,    .384)
    (.7,    .362)
    (.8,    .323)
    };
    \addlegendentry{Baseline Model}
\addplot[
    color=gray,
    mark=triangle,
    mark size=0pt,
    dashed,
    ]
    coordinates {
    (.2,    .385)
    (.8,	.385)
    };
\addplot[
    color=gray,
    mark=triangle,
    mark size=0pt,
    dashed,
    ]
    coordinates {
    (.2,    .501)
    (.8,	.501)
    };
\addplot[
    color=gray,
    mark=triangle,
    mark size=0pt,
    dashed,
    ]
    coordinates {
    (.2,    .539)
    (.8,	.539)
    };
\addplot[
    color=gray,
    mark=triangle,
    mark size=0pt,
    dashed,
    ]
    coordinates {
    (.2,    .433)
    (.8,	.433)
    };
\addplot[
    color=gray,
    mark=triangle,
    mark size=0pt,
    dashed,
    ]
    coordinates {
    (.2,    .519)
    (.8,	.519)
    };
\addplot[
    color=gray,
    mark=triangle,
    mark size=0pt,
    dashed,
    ]
    coordinates {
    (.2,    .251)
    (.8,	.251)
    };
\end{axis}
\end{tikzpicture}
\caption{
Performance in terms of Cohen's $\kappa$ \cite{Cohen1960} of our curriculum-learning and baseline models compared with that of pathologists annotators.
The $x$-axis shows different confidence thresholds for our models, and the $y$-axis displays the average agreement with each pathologist in terms of Cohen's $\kappa$.
The 4.5\% improvement in AUC (Table \ref{tab:main_table}) of the curriculum-learning model compared with the baseline model translated to an .089 improvement in Cohen's $\kappa$, allowing the curriculum-learning model to achieve agreement on par with the pathologist mean.
}
\vspace{-3mm}
\label{fig:cohens-kappa}
\end{centering}
\end{figure}

\vspace{1mm}
\section{Discussion}
\label{sec:discussion}
\noindent \textbf{Human and machine notions of difficulty.} 
Our study has presented a transparent analysis of the requirements of curriculum learning, proposing that histopathology image analysis tasks present a range of difficulty among examples and that readily-available annotator agreement can be used as a natural proxy for ranking images. 
Experimentally, we found that using this natural proxy as a curriculum to train classifiers can yield significant performance improvements. 

Some prior work has demonstrated that what makes an image difficult for neural networks to classify might not always match what makes it difficult for human annotators, an observation that recent work on adversarial examples takes advantage of \cite{Szegedy2014}.
Our study explores a converse idea for histopathology image classification, finding that machine notions of difficulty do correlate with human annotator agreement and contending that annotator agreement can be a useful proxy for facilitating curriculum learning.

\vspace{1.2mm} \noindent \textbf{Implications on medical image analysis.}
Our work also has implications for how labels are used in the histopathology image analysis and medical image analysis domains.
Much prior work in these domains has used majority voting or senior pathologist resolution to resolve labels, only retaining a single label for each image without distinguishing between images with high annotator and low annotator agreement.
Our method could instead leverage the annotation agreement for curriculum learning. 
In Table \ref{fig:agreement_examples} in the supplementary materials section, we list past work for which multiple annotator agreement levels appears to be available and therefore our method could be applicable (although most are private datasets).

Certainly, not every scenario in medical imaging is ideal for applying curriculum learning. 
For instance, both pathologists and deep learning models have achieved high performance in distinguishing high-grade lung cancers and normal tissue, which is considered a relatively easy task that is less likely to exhibit a range of difficulty among examples.
Another negative example could be detecting bone fractures: since bone fractures are typically caused by instantaneous traumatic events, there is no progression of disease development, so we are less certain that a range of example difficulty exists.
Potential scenarios conducive to curriculum learning by annotator agreement should ideally involve a progression of disease development and be challenging problems where even specialized pathologists might disagree.
In particular, we believe that cancer datasets can benefit from curriculum learning because of the inherent natural progression of cancers (i.e., cancer develops over time and is not sudden).
This could include tasks such as distinguishing among subtypes of lung adenocarcinoma or assessing tumor proliferation in breast cancer tissue samples.

\vspace{1.2mm} \noindent \textbf{Limitations.}
Though we intentionally addressed a common, clinically-important, and diagnostically-challenging problem of colorectal polyp classification for empirical evaluation---the best dataset that we were able to collect and annotate with multiple annotators at this time---our study nonetheless only uses a single dataset.  
As such, although our approach seems effective in its current form, we consider our results as an invitation for further exploration in this direction rather than a validation of curriculum learning for all histopathology classification tasks.

Furthermore, our dataset contained annotations from seven pathologist annotators, allowing us to categorize images into four discrete levels of difficulty. 
For medical image datasets that have fewer annotators and therefore fewer categorized levels of annotation difficulty, we suspect that the benefits of our approach could be slightly diminished.
For example, our coarse annotator agreement method in $\S$\ref{sec:proxies}, which only used two levels of annotator agreement, achieved a smaller performance improvement (1.5\%) than our four-level annotator agreement method (4.5\%).
Moreover, our dataset size is modest in the medical imaging domain, and so whether these curriculum methods work for other dataset sizes is still relatively unexplored.
We believe, however, that our curriculum learning methodology might still be worth exploring for these datasets with fewer annotators, as even small improvements in performance are important given the high cost of data annotation and the importance of accuracy toward patient outcomes.

\section{Further Related Work and Conclusions}
\label{sec:further-related-work}
Although we argue that histopathology imaging is a promising context for curriculum learning, we are not the first to explore curriculum learning in the medical imaging domains.
Oksuz et al.\ \cite{Oksuz2019} used curriculum learning for artifact detection by first training on heavily-corrupted images and then introducing less-corrupted images, thereby improving performance on borderline cases.
Maicas et al.\ \cite{Maicas2018} used a meta-training approach called teacher-student curriculum learning to improve breast screening classification on a weakly-labeled dataset.
Moreover, Jesson et al.\ \cite{Jesson2017} used adaptive curriculum sampling to better detect lung nodules in extreme class-imbalance scenarios, and
Oksuz et al.\ \cite{Oksuz2019} applied a curriculum based on disease severity levels in radiology reports (e.g., mild, moderate, severe).

These prior studies have demonstrated empirical evaluation of curriculum learning and helped inspire our work, but their methods tend to be complicated and use inductive biases specific to certain datasets.
For instance, the approach of Oksuz et al.\ \cite{Oksuz2019} only applies to artifact detection, the approach of Jesson et al.\ \cite{Jesson2017} is specific to segmentation tasks, and disease-severity information from radiology reports used in Oksuz et al.\ \cite{Oksuz2019} might not exist for many datasets and can be challenging to parse.
Our method, on the other hand, is easy to implement, presents no modifications to network architecture, and only requires annotator agreement data, which is often readily available. 

Based on a thoughtful analysis of the assumptions in curriculum learning, we have presented a simple yet effective curriculum learning framework which leverages easily-obtained annotator agreement data. 
In histopathology image analysis, where data collection and annotation can be especially costly, it is important to combine the natural properties of classification tasks with the most-appropriate inductive biases.
We aim to have provided a well-motivated argument for more intentional application of curriculum learning to readers from both computer vision and medical imaging analysis backgrounds.

\newpage
{\small
\bibliographystyle{splncs04}
\bibliography{egbib}
}

\newpage
\begin{figure*}[h!]
    \centering
    \includegraphics[width=\linewidth]{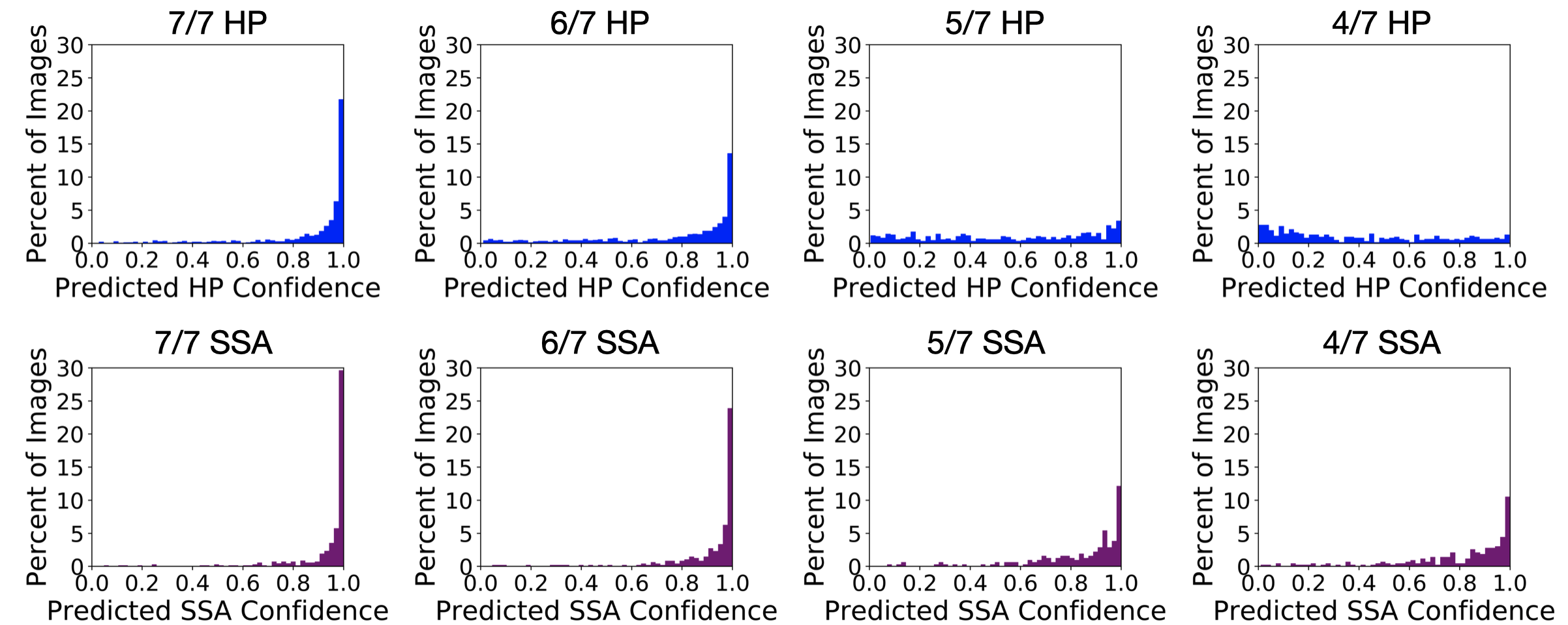}
    \vspace{-1.3mm}
    \caption{Distribution of predicted confidences by a pre-trained model fine-tuned on our dataset for different annotator agreement levels (indicated above each plot).}
    \label{fig:agreement_distributions}
\end{figure*}
\begin{table*}[h]
    \centering
    \begin{tabular}{l | c | c}
        \toprule
        Dataset & Annotators & Resolution Method\\
        \midrule
        Head CT Scans \cite{Chilamkurthy2018} & 3 & Majority Vote\\
        Lymph Node Metastases \cite{Bejnordi2017} & 3 & Senior Expert Verification\\
        Diabetic Retinopathy \cite{Gulshan2016} & 7 & Majority Vote\\
        Chest Radiograph \cite{Irvin2019} & 3 & Majority Vote\\
        Lung Carcinoma \cite{Kanavati2020} & 3 & Senior Expert Verification\\
        Follicular Lymphoma Grading \cite{Sertel2008} & 5 & Majority Vote\\
        Ulcer Recognition \cite{Wang2019} & 3 & Senior Expert Verification\\
        Colorectal Polyps \cite{Wei2020} & 5 & Majority Vote \\
        Breast Cancer \cite{Zhou2019} & 3 & Senior Expert Verification\\
        \bottomrule
    \end{tabular}
    \vspace{1mm}
    \caption{Examples of image classification tasks from prior work where annotator agreement is accessible in principle.}
    \label{tab:annotator_agreement_uses}
\end{table*}
\section*{Supplementary Materials}\label{apd:first}
Figure \ref{fig:agreement_distributions} on the next page shows the correlation between difficulty as perceived by a pre-trained model and pathologist annotators.
Table \ref{fig:agreement_examples} on the next page lists examples of medical image classification tasks from prior work where we believe annotator agreement data to be accessible.

\end{document}